# Wide Open Gazes: Quantifying Visual Exploratory Behavior in Soccer with Pose Enhanced Positional Data

Joris Bekkers

U.S. Soccer Federation, Atlanta, USA    UnravelSports, Breda, Netherlands

## 1. Abstract

Traditional approaches to measuring visual exploratory behavior in soccer rely on counting visual exploratory actions (VEAs) based on rapid head movements exceeding 125°/s, but this method suffer from player position bias (i.e., a focus on central midfielders), annotation challenges, binary measurement constraints (i.e., a player is scanning, or not), lack the power to predict relevant short-term in-game future success, and are incompatible with fundamental soccer analytics models such as pitch control. This research introduces a novel formulaic continuous stochastic vision layer to quantify players' visual perception from pose-enhanced spatiotemporal tracking. Our probabilistic field-of-view and occlusion models incorporate head and shoulder rotation angles to create speed-dependent vision maps for individual players in a two-dimensional top-down plane. We combine these vision maps with pitch control and pitch value surfaces to analyze the awaiting phase (when a player is awaiting the ball to arrive after a pass for a teammate) and their subsequent on-ball phase. We demonstrate that aggregated visual metrics - such as the percentage of defended area observed while awaiting a pass - are predictive of controlled pitch value gained at the end of dribbling actions using 32 games of synchronized pose-enhanced tracking data and on-ball event data from the 2024 Copa América. This methodology works regardless of player position, eliminates manual annotation requirements, and provides continuous measurements that seamlessly integrate into existing soccer analytics frameworks. To further support the integration with existing soccer analytics frameworks we open-source the tools required to make these calculations.

## 2. Introduction

Soccer analytics has evolved rapidly over the past decade, with advances occurring across multiple different domains. Event-level data research has enabled the valuation of players and their actions by assigning probabilities to on-ball events based on the likelihood they lead to a goal [15], distinguishing of soccer-specific player roles [1], identification of playstyles from passing patterns using graph theory [8], and the quantification of pass creativity [35]. Simultaneously, sophisticated tracking data (as described by [3]) - which capture continuous player and ball movement multiple times per second - have enabled risk-reward assessment of passes [32], reinforcement learning frameworks for optimal decision-making [33], and evaluation of off-ball players by comparing their movements with predicted trajectories [41]. This data has further enabled: formation analysis across distinct phases of play [4], while also enabling advanced defensive analytics including frameworks for quantifying pressing intensity [5], evaluating individual player contributions within pressing situations [25], and modeling the success of counterattacks [9]. Foundational work in the area of spatial analysis [40] ultimately evolved into sophisticated pitch control models [37, 20] - which quantify the probability of each team controlling different areas of the pitch at any given moment, and a complementary pitch value framework that assign value to strategically important locations [20]. The explosive growth of research, data adoption, and data availability in soccer has even led to the creation of a standardized format for soccer data [2].

Technological advances in computer vision and human pose estimation from the likes of OpenPose [10], HRNet [39] and HigherHRNet [14] have ushered in the next frontier in soccer analytics. These advances have enabled the extraction of detailed body pose information, including head and shoulder positioning, allowing us to steer our focus towards the incorporation of visual perception into our soccer modeling.

Traditional approaches to understanding visual perception in soccer have relied heavily on controlled laboratory settings using eye-tracking technology to examine gaze behaviors and visual search patterns [11, 36, 41, 30]. While these studies have provided valuable insights into the perceptual-cognitive processes underlying expert performance - demonstrating that skilled players employ more efficient visual search strategies with shorter fixation durations across more relevant locations [36, 42] - they are fundamentally limited by their experimental conditions. These laboratory studies typically require participants to watch video footage and make decisions about hypothetical game situations rather than measuring visual behavior during actual match play [29]. A systematic review revealed that no studies have investigated exploration behaviors during actual open-play situations, with most research not reflecting the complex, dynamic demands of real soccer performance [29]. While another review, covering 2016 through 2022, shows encouraging progress toward more realistic research conditions and a higher emphasis on natural motor responses. However, these advances have been accompanied by persistent methodological limitations, including smaller sample sizes, outdated eye-tracking technology, and insufficient data quality reporting standards [24].

The translation of these laboratory findings to on-field measurement has primarily relied on counting *visual exploratory actions* (VEAs) based on rapid head movements exceeding 125°/s [22]. However, this approach exhibits several fundamental limitations that constrain its applicability to performance analysis. VEAs demonstrates an inherent bias towards playing positions that require frequent visual exploratory actions (e.g., central midfielders) [16, 22, 23, 28], are challenging to annotate accurately with substantial inter-observer variability [26, 13, 23], prove difficult to collect reliably from standard 25 FPS tracking data [26], and rely on binary measurements (i.e. rapid head movement, or not) that fail to accommodate data inaccuracies or capture the continuous nature of visual attention [26, 13]. Furthermore, studies validating VEAs as a performance metric rely on simplistic in-game measures like pass success or forward pass success [23, 28]. These associations are not even statistically significant for some position groups, undermining the metric's utility [23].

Additionally, VEA frequency varies significantly based on contextual factors such as pitch position, phase of play, and opponent pressure [27, 31]. Furthermore, a recent study [12] found no significant differences in VEA between super elite and elite players, questioning the fundamental premise that VEA frequency serves as a reliable performance differentiator.

This research introduces a novel integration of pose estimation data and spatial modeling to address the limitations of traditional VEA measurement. We introduce a *vision* layer that utilizes positional tracking data enhanced with head and shoulder rotation angles extracted from this estimated body pose data. We use this stochastic vision layer - which represents the probability distribution of where each player is directing their visual attention across the pitch - to quantify *scanning* behavior continuously for each player individually, rather than simply counting rapid head movements.

By combining our vision layer with established *pitch control* and *pitch value* frameworks [20], we demonstrate that modeling this stochastic vision layer can be highly valuable for quantifying visual exploratory behavior. Even aggregated visual metrics (such as percentage of defensive area observed while awaiting a pass from a teammate) are strong predictors of pitch value gained (or lost) at the end of a player's subsequent dribble.

Our approach addresses the fundamental limitations of contemporary VEA research by establishing a formulaic method that works regardless of player position, eliminates reliance on manual data annotation, and provides continuous rather than binary measurements. By defining visual exploratory behavior as a two-dimensional stochastic grid spanning the field of play, we create direct compatibility with existing analytical frameworks such as *pitch control* and SoccerMap [19], enabling seamless integration into current methodologies. Additionally, we demonstrate that traditional VEA counting methods - captured at 25 FPS - lack predictive power for subsequent on-ball performance outcomes, highlighting the need for more sophisticated approaches to quantifying visual perception in soccer and further questioning the reliability of VEAs as performance indicators.

## 2.1. Research

We start by modeling the vision layer for an individual player (detailed in Section 3.1). This consists of two components: a probabilistic **field of view map**, and a probabilistic **occlusion map** created by all other players on the field as perceived from the perspective of this individual player (Section 3.1). To encourage reproducibility and to support more research into visual exploratory behavior, we share code and data to construct these maps on the U.S. Soccer Federation GitHub.[1]

We combine these maps - using element-wise matrix multiplication - with two components from Fernández & Bornn (2018) [20]: a probabilistic *pitch control* map that approximates the controlled space by both attacking and defending teams, and a *pitch value* map that dynamically describes the value of each field segment learned from defensive team setups given the ball location. We additionally introduce a modification parameter to the pitch control surface calculation to reduce the amount of space controlled by a player. We use this as a proxy for the locations on the pitch a player can reach within a very short period of time, which we call **imminent pitch control**. Combining these facets gives us the ability to measure visual exploratory behavior during a match for any player, at any moment in a game, and directly relate it to pitch value gained or lost.

To validate our methods we construct a set of Binary Classification XGBoost models to learn if the modeled visual exploratory behavior is predictive of short-term in-game future success, measured as a significant increase or decrease of controlled pitch value at the start and end of a phase (see

---

[1] https://github.com/USSoccerFederation/ssac26_visual_exploratory_behavior

Section 4). We specifically focus on the *awaiting* phase and the subsequent *on-ball* phase. The awaiting phase is defined as the moments when a player is awaiting the ball to arrive after their teammate has executed a pass and the subsequent *on ball* phase is simply the on-ball movement (e.g., dribble) the player executes after receiving the ball. The combination of vision and pitch value gives us a comprehensive approach to model individual players' ability to visually identify valuable space and act accordingly in key moments of a team's possession chain.

**2.2. Validation Dataset**

To validate this research, we use 32 games of broadcast tracking data provided by Respo.Vision [34] to all teams participating in the 2024 Copa América. This data (recorded at 25 frames per second) contains team and player identifiers, and player and ball $x$ and $y$ coordinates. It is enhanced with head rotation, shoulder rotation, and hip rotation angles of each player in the two-dimensional plane. Figure 1 shows a two-dimensional, top-down comparison of a snapshot of regular tracking data compared to its limb tracking enhanced counterpart.

To complement this data, and to extract the relevant phases (*awaiting* and *on-ball*) the positional data is enhanced with synchronized event data from StatsPerform (Opta) [38] using a combination of *kloppy* [44], *socceraction* [15] and a simplified version of the *ETSY* rule-based synchronization algorithm [43]. Ball receptions are identified as the first local minimum of the distance between the player receiving the ball and the ball.

In total, our dataset consists of approximately 2.3 million frames of tracking data and over 60,000 events. Each event is assigned to a single frame of tracking data. We filter this data to include only uncontested sequences of open play, defined as a sequence of play that has at least 1 successful pass and as any action that happens at least 7 seconds after the most recent set-piece. This yields approximately 14,000 awaiting moments with subsequent on-ball events.

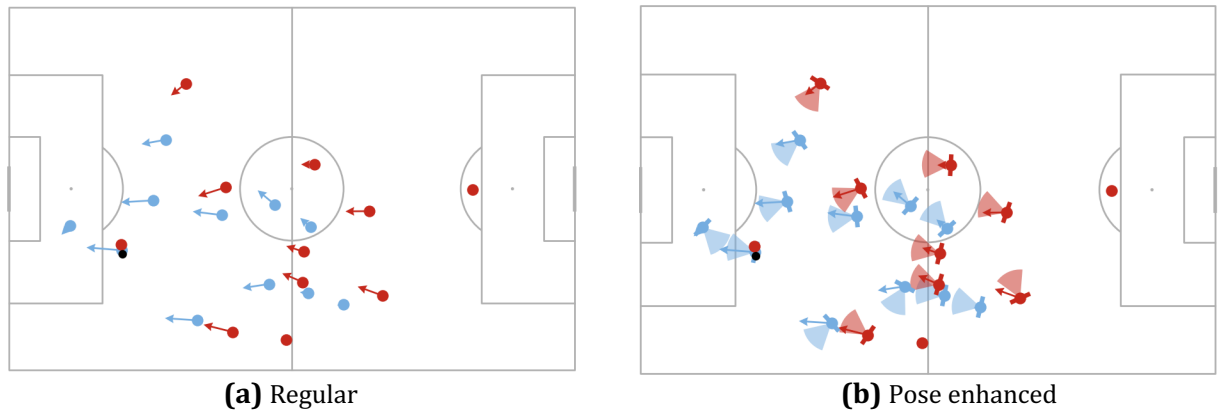

**(a)** Regular        **(b)** Pose enhanced

**Fig.1.** Snapshots of Respo.Vision broadcast tracking data from Argentina v Canada in the 2024 Copa América excluding (a) and including (b) head and shoulder orientation.

Missing angles (head, shoulder, or hip) are interpolated per player using Euler's formula, which decomposes angles into real and imaginary components that are interpolated independently and then reconstructed into angles. This approach is necessary because angles are cyclical, requiring interpolation along the shortest angular path.

# 3. Methodology

In this section, we illuminate all aspects required to construct a comprehensive vision map to superimpose on pitch control and pitch value models. As a visual supplement to this section five short videos are made available on [7] and [2]. Additionally, a reference table defining all mathematical notation used in this research can be found in Appendix A.

### 3.1. Vision

To build a comprehensive two-dimensional, top-down representation of a player's visual perception, we create two complementary models: a **vision** model (see Section 3.1.1) to quantify a player's ability to see other players on the pitch, and an **occlusion** model (see Section 3.1.2) to describe any player on the pitch blocking (parts) of a player's vision. Together, these two facets will give us the complete vision map of each player expressed as a 105x68 grid (i.e., the length and width of the pitch).

Because human vision is not perfect, and objects in the peripheral vision, or those at a greater distance, are more difficult to locate exactly, especially at high speeds, we introduce a probabilistic method to quantify both the vision and occlusion models. We use head orientation as a proxy for eye orientation, as they are often oriented in the same direction as reported by [18, 13, 26]. Approximately 92% of gaze shifts are in the same direction as head movement within the horizontal plane [18].

### 3.1.1 Field of View

To start, we create a map of the *binary field of view* ($V_\beta$) for player *i*, as shown in Figure 2a. We assume that binocular vision spans 120° [21] and we assume finite vision toward the edges of the pitch.

Subsequently, we introduce a probabilistic layer on top of this binary field of view that should more closely resemble how accurately a player can identify the location of other players given the inherent uncertainty. We denote this as the *field of view* ($V_\rho$) of player *i* at location *p*, at time *t* with head angle $\theta_h$ and speed *v* (see Formula 1).

$$V_\rho(p, t, \theta_h, v)_i = R\big(c_r(v_i)\big) \odot A\big(c_a(v_i)\big) \odot V_\beta(p, t, \theta_h)_i \qquad (1)$$

$$R\big(c_r(v_i)\big) = exp\left[-c_r \left(\frac{d}{\sigma_r}\right)^2\right] \qquad (2)$$

$$A\big(c_a(v_i)\big) = exp\left[-c_a \left(\frac{\theta_a}{\sigma_a}\right)^2\right] \qquad (3)$$

---

[2] https://unravelsports.com/ssac/26/veb.html

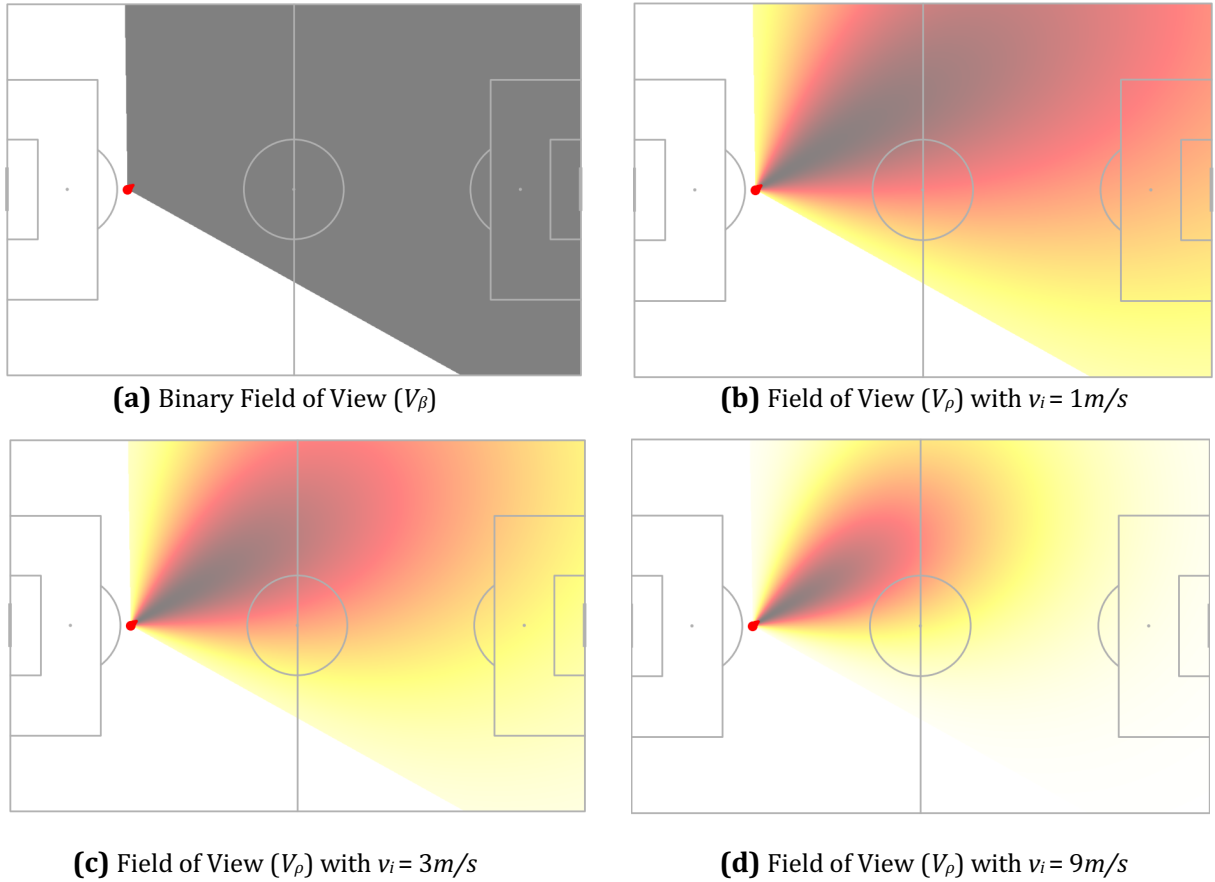

**(a)** Binary Field of View ($V_\beta$)  
**(b)** Field of View ($V_\rho$) with $v_i = 1 m/s$  
**(c)** Field of View ($V_\rho$) with $v_i = 3 m/s$  
**(d)** Field of View ($V_\rho$) with $v_i = 9 m/s$  

**Fig.2.** An individual player $i$ at location (22, 34) with head angle 31°

We introduce $R$ and $A$ (Formula 2 and 3) to describe the vision of player $i$ as a Gaussian that decreases as the observed location moves further away from player $i$ radially and angularly. Both

$R$ and $A$ are directly influenced by the players' speed ($v_i$) - because we assume higher speeds result in a loss of spatial awareness and a narrower focus on a player's immediate surroundings - through scaling parameters $c_r$ and $c_a$ respectively (see Appendix B). Here, $c_r$ determines the decay rate of the Guassian, $d$ denotes the distance away from player $i$, and constant $\sigma_r$ is the standard deviation of the depth of vision. $A$ in turn describes a Gaussian that explains the width of vision, where $c_a$ is a scaling parameter controlling the decay of vision as objects move further into the peripheral vision of player $i$, $\theta_a$ denotes the angle away from the focal point and constant $\sigma_a$ is the standard deviation of the angular vision.

Examples of $V_\rho$ with different values for speed $v$ can be found in Figure 2b, 2c and 2d.

### 3.1.2 Occlusions
Player $i$ is not the only player occupying the pitch, and as a result, those other players $J$ might obstruct player $i$'s vision. Therefor we enhance the veracity of the field of view map ($V_\rho$) of player $i$ by incorporating visual obstructions caused by other players on the field in the form of an occlusion map ($V_\Phi$). $V_\Phi$ can be understood as the combination of every individual occlusion map ($V_{\phi,i,j}$) of player $j$ blocking parts of player $i$'s vision. The size of the occlusion caused by player $j$ is influenced by the inter-player distance and the shoulder rotation of player $j$. This shoulder rotation ($\theta_s$) is used to determine how much of an obstruction player $j$ forms given their rotation towards player $i$. If a player is angled sideways from the perspective of player $i$ they occupy less space in player $i$'s field of vision compared to looking directly at their torso.

An occlusion map $V_{\phi,i,j}$ depends on the head angle ($\theta_h$) and location of player $i$ (now explicitly denoted as $p_i$), and the location ($p_j$) and shoulder angle ($\theta_s$) of player $j$.

To build a single occlusion map ($V_{\phi,i,j}$) describing the obstructed areas caused by player $j$ as observed by player $i$ we create a probabilistic *ray* ($Q_{i,j}$) projected from player $i$ through player $j$ that follows Formula 5, and is depicted in Figure 3a. We control the maximum probability of obstruction with parameter $\alpha$.

Additionally, we construct a binary mask $V_{o,i,j}$ that models the unobstructed view between players $i$ and $j$ (see Figure 3b). Multiplying these three components yields occlusion map $V_{\phi,i,j}$ as described in Formula 4 and depicted in Figure 3c. Thus, this models the space that player $i$ is (partially) unable to observe as a result of player $j$.

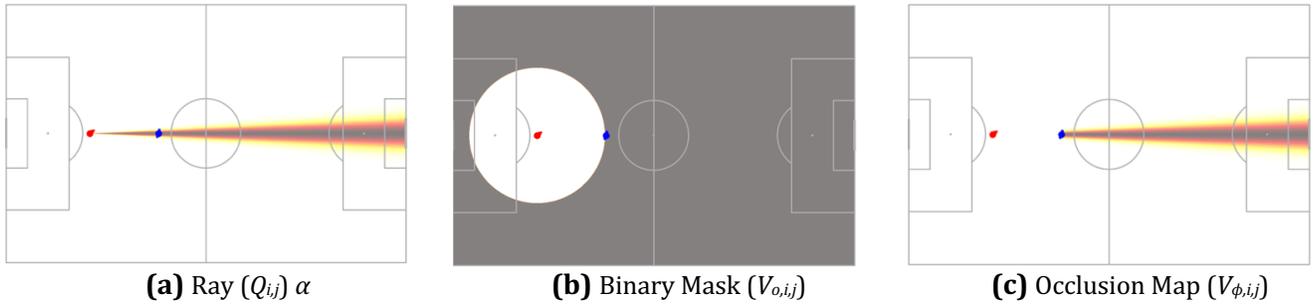

**(a)** Ray ($Q_{i,j}$) $\alpha$     **(b)** Binary Mask ($V_{o,i,j}$)     **(c)** Occlusion Map ($V_{\phi,i,j}$)

**Fig.3.** An individual occlusion map ($V_{\phi,i,j}$) of player $j$ as perceived by player $i$.

$$V_{\phi,i,j}(p_i, \theta_h, p_j, \theta_s, t) = Q_{i,j}(p_i, p_j, \theta_s, t) \odot V_o(p_i, \theta_h, p_j, t) \, \alpha \quad (4)$$

$Q_{i,j}$ uses the same angular approach as $A$ (in Formula 3) with $\theta_q$ and $\sigma_q$ mimicking the functionality of $\theta_a$ and $\sigma_a$, respectively. However, now the scaling parameter $c_q$ (Formula 6) is determined by the apparent angular width of player $j$ from the perspective of player $i$, considering the shoulder rotation of player $j$ ($\theta_s$). Here $\delta_{i,j}$ is the distance between players $i$ and $j$, and $\omega_a$ is the apparent angular width of player $j$ as perceived by player $i$ (derived from shoulder width $\omega_s$ and torso depth $d_s$. See Appendix C for a complete derivation).

$$Q_{i,j}(p_i, p_j, \theta_s, t) = \exp\left[-c_q \left(\frac{\theta_q}{\sigma_q}\right)^2\right] \quad (5)$$

$$c_q(p_i, p_j, \theta_s, t) = \frac{\delta_{i,j}}{\omega_\alpha} \tag{6}$$

In essence, this means that $\omega_\alpha$ models how much of an obstruction player *j* forms given their rotation towards player *i*.

Parameter $\delta_{i,j}$ ensures the apparent width of player *j* is scaled with distance, because objects that appear closer to player *i* should cast a larger shadow than those at a greater distance. This concept is illustrated by three examples in Figure 4.

$$V_{\Phi,i} = \prod_{j \in J}(1 - V_{\phi,i,j}) \tag{7}$$

Finally, we model $V_\Phi$ as the element-wise product of all occlusion matrices for player *i* observing all players *J*, as shown in Formula 7 and in Figure 5.

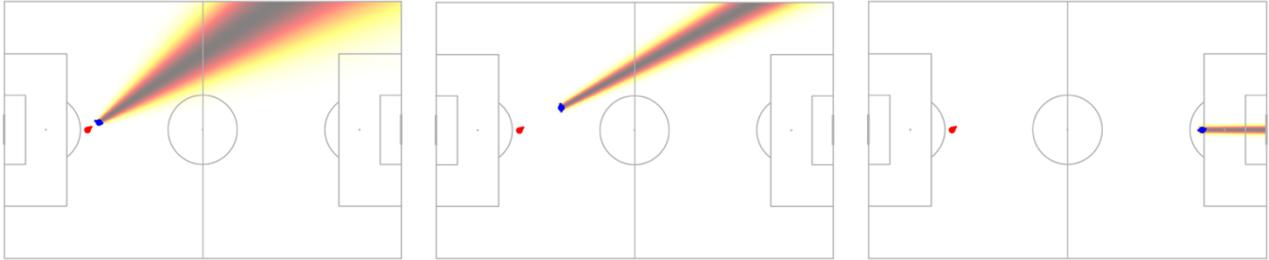

**Fig.4.** Three examples of occlusion maps to demonstrate how the ray $Q_{i,j}$ is appropriately controlled by scaling parameter $c_q$ to never be wider than necessary.

### 3.1.3 Complete Vision Map of Player *i*
We combine field of view map ($V_{\rho,i}$) and occlusion map ($V_{\Phi,i}$) through elementwise matrix multiplication to obtain the total vision map ($V$) for player *i* following Formula 8.

$$V_i = V_{\Phi,i} \odot V_{\rho,i} \tag{8}$$

Figure 7 depicts two examples of the total vision map at two different speeds ($v_i$ = 1*m/s* and $v_i$ = 9*m/s*) and Figure 6 depicts Argentina player Rodrigo de Paul observing players directly after receiving the ball in a moment during the Copa América final. Five videos of this sequence of play are shared at [7] and [2].

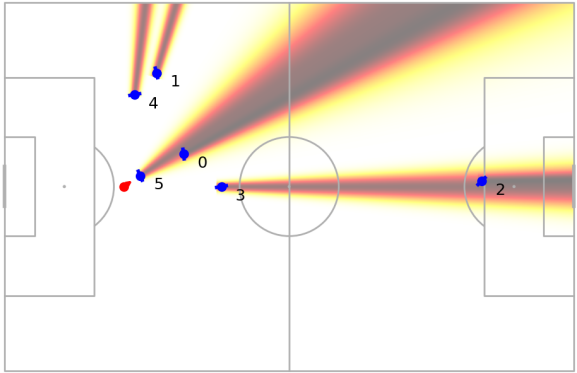

**Fig.5.** Combined occlusion map (shown as $1 - V_{\Phi,i}$) where player $i$ (red) observes 6 players.

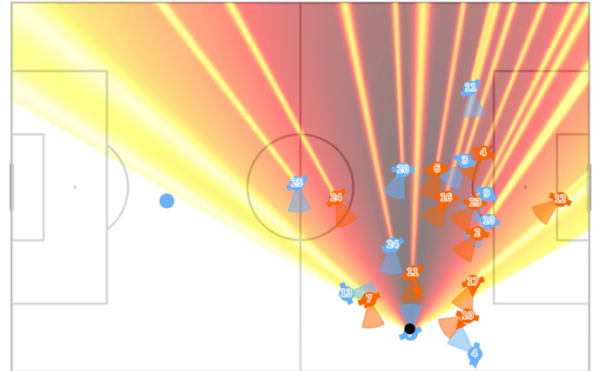

**Fig.6.** Rodrigo De Paul observing players in the Copa América final [7].

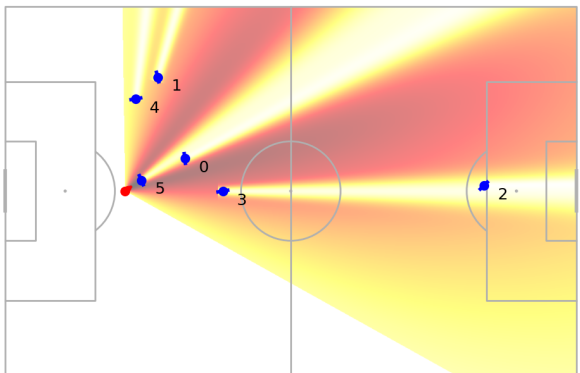

**(a)** Vision Map ($V_i$) with $v_i = 1 m/s$

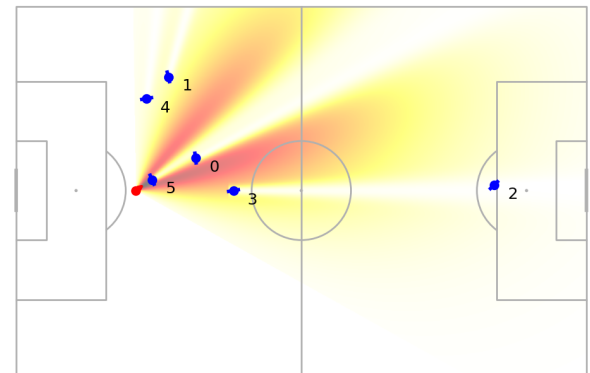

**(b)** Vision Map ($V_i$) with $v_i = 9 m/s$

**Fig.7.** Two examples of the final vision map ($V_i$) with different player speeds ($v_i$).

## 3.2. Pitch Control

Now that we have laid out the vision model, we model the quantity and quality of occupied space for each individual player. We accomplish this by relying on two concepts from Fernández & Bornn (2018) [20], namely *pitch control*; the space most likely controlled by either team (or player) at a given moment, and *pitch value*; the value of the space on the pitch.

### 3.2.1 Imminent Pitch Control

Fernández & Bornn introduce their *pitch control* (PC) framework to quantify the probability that a player controls a location on the field at any moment given their current location, speed, magnitude and distance to the ball by assigning a scaled and rotated bi-normal distribution to determine which player could realistically reach an area first. Figure 8 depicts their original model applied to a single frame of tracking data.

We use their approach and introduce a scaling parameter ($c_{in}$) to the player influence radius formula ($R_i$, see [20]) to decrease the control radius of each player (see Figure 9). This operation allows us to mimic the space a player could move towards in a short amount of time (**imminent** **pitch control**). For simplicity, we denote this as $PC^{C_{in}}$ (e.g., $PC^{0.5}$ for a smaller surface, as shown in Figure 9).

Furthermore, we define the defending team (Figure 10a), the attacking team (Figure 10b) or all players except player *i* (Figure 10c) to determine the imminent space controlled by each entity individually.

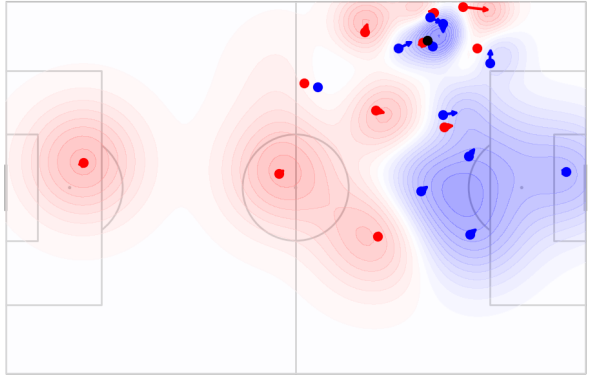
**Fig.8.** Default Pitch Control

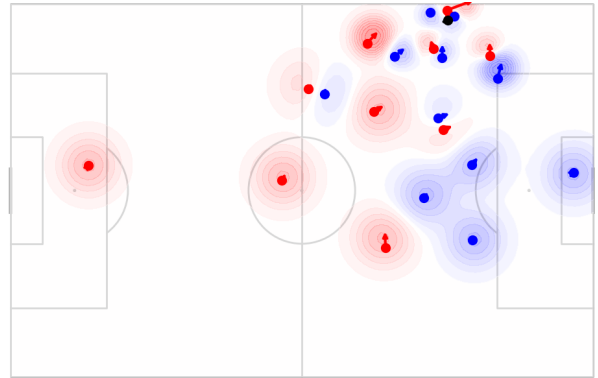
**Fig.9.** "Imminent" Pitch Control ($c_{in}$=0.5)

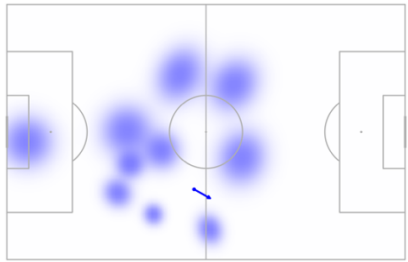
**(a)** Attacking team excl. player *i* ($PC^{0.5}_{j_{att}}$ where $i \notin J$)

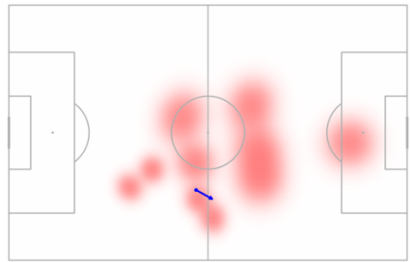
**(b)** Defending team ($PC^{0.5}_{j_{def}}$)

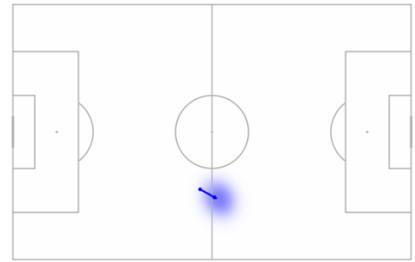
**(c)** Attacking player *i* ($PC^{0.5}_i$)

**Fig.10.** Imminent pitch control for player *i* depicted by the blue arrow, teammates (blue) and opponents (red).

### 3.2.2 Pitch Value
Fernández & Bornn additionally introduce a feed forward neural network trained to estimate the pitch influence of the defending team given the location of the ball. They use this to estimate the location of valuable attacking space on the pitch, under the assumption that on average the defensive team positions itself in relation to the ball to cover the most valuable spaces. To obtain this *pitch value* ($V_l$) we train a similar feed forward neural network on pairs of ball location and associated defensive influence surfaces for a random subset (n=19,504, less than 1%) of open play moments from our Copa América dataset (see Figure 11a for an example prediction).

$$\widehat{V_l} = V_l \odot V_\eta \qquad (9)$$

In Figure 11b we depict the *distance to goal pitch normalization surface* ($V_\eta$) as proposed by [20]. This captures the intuitive understanding that space closer to the opponent's goal is more valuable. Figure 11c depicts the normalized pitch value given ball location $p_b$ which is computed using element-wise matrix multiplication following Formula 9.

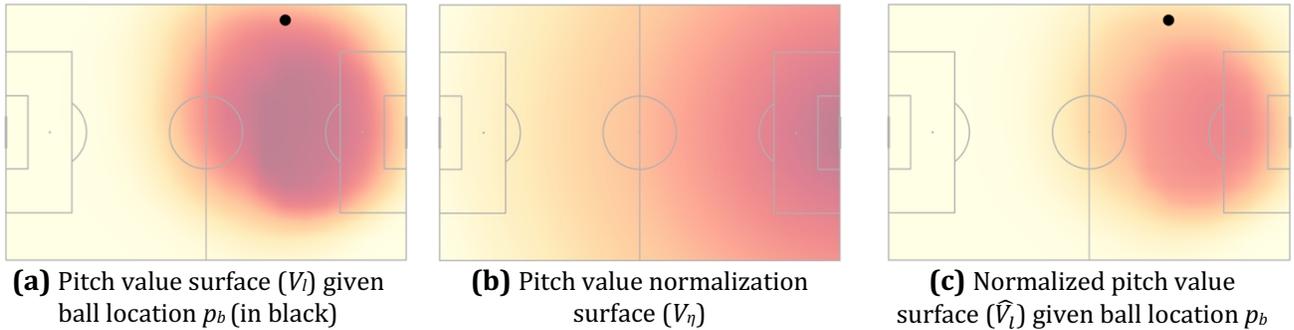

**(a)** Pitch value surface ($V_l$) given ball location $p_b$ (in black)   **(b)** Pitch value normalization surface ($V_\eta$)   **(c)** Normalized pitch value surface ($\widehat{V_l}$) given ball location $p_b$

**Fig.11.** Predicted pitch values in a range between 0 and 1.

## 4. Validation

We now evaluate our vision model by conducting an ablation study to assess whether seeing certain spaces (e.g., observed imminent attacking pitch control, or observed imminent defending pitch control, see Figure 12) during the *awaiting* phase leads to players subsequently reaching more valuable space at the end of their *on-ball* phase. The dataset spans all relevant frames ($n$=171,318) of our 14,000 awaiting moments aligned with the final on-ball frame after this awaiting phase.

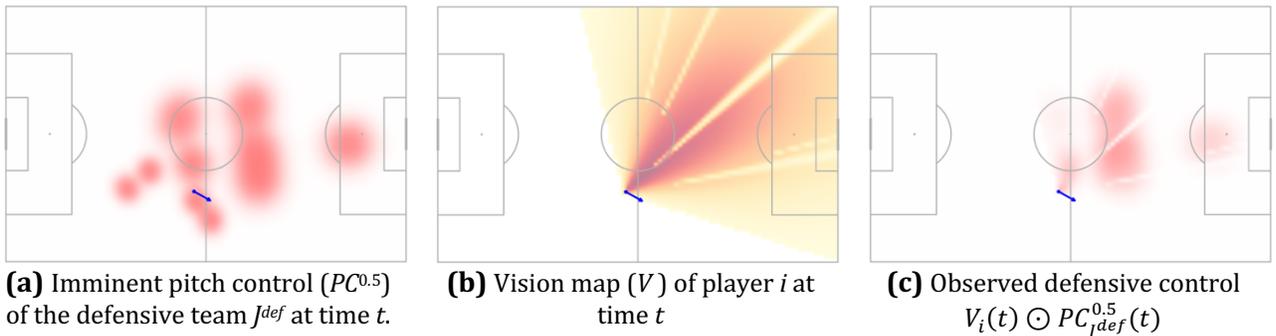

**(a)** Imminent pitch control ($PC^{0.5}$) of the defensive team $J^{def}$ at time $t$.   **(b)** Vision map ($V$) of player $i$ at time $t$   **(c)** Observed defensive control $V_i(t) \odot PC^{0.5}_{J^{def}}(t)$

**Fig.12.** An example of combining multiple models.

We construct four Binary Classification XGBoost models (see Section 4.2) trained on balanced datasets with standardized features. The binary labels (see Section 4.1) describe whether player $i$'s pitch value at time $t$ has significantly increased or decreased at the end of the on-ball phase (at time $t^{end}_{on\_ball}$).

### 4.1. Labels

The increase in pitch value is measured as the ratio ($p_{rat}$) of the instantaneous pitch value ($P_v(t)$) of the space occupied by player $i$ during the awaiting phase at current time $t$ and the instantaneous pitch value occupied by the same player during the end of the subsequent phase ($t_{on\_ball}^{end}$), as shown in Formula 11. In other words, assume a player at time $t$ ("now") is *awaiting* a ball arrival, if they controlled significantly more imminent pitch value at the end of the subsequent on-ball phase this sample is annotated with a positive label.

$$p_{rat}(t) = \frac{P_v(t_{on\_ball}^{end})}{P_v(t) + P_v(t_{on\_ball}^{end})} \tag{11}$$

We now apply these binary labels to each sample following Formula 12. Here $y = 0$ constitutes a clear decrease in pitch value and $y = 1$ constitutes a clear increase in pitch value. We choose to omit samples with a $p_{rat}$ between 0.35 and 0.65, because these moments introduce significant noise.

$$y = \begin{cases} 0 & if \quad p_{rat} < 0.35 \\ Excl. & if \quad 0.35 \leq p_{rat} \leq 0.65 \\ 1 & if \quad p_{rat} < 0.65 \end{cases} \tag{12}$$

### 4.2. Ablation Models

The ablation study considers four models, built on four datasets. By progressively adding more features to subsequent datasets we can quantify the specific contribution of our vision-based features to the overall model performance. Each model employs 150 estimators, a learning rate of 0.05, maximum depth of 4, and early stopping to prevent overfitting. The datasets are split into test and train sets by game to ensure no leakage occurs between sets. Below is an outline of the four model configurations.

- **Baseline model** includes only distance to the center of the goal.
- **Traditional VAE model** adds the total number of regular VEAs (when the angular velocity of the players' head exceeds 125°/$s$) in the past second [26, 13], and in the past two seconds, and between $t_{await}^{start}$ and time $t$.
- **Regular model** further adds distance to the goal line; distance to the center of the pitch in the $x$ and $y$ directions; the components of the velocity vectors ($vx$ and $vy$); and standardized position labels derived using Elastic Formation and Position Identification (EFPI) [6]. After assigning the position labels we simplify them into one of six general position labels, namely Wide Midfielder, Wide Back, Center Forward, Central Midfielder, Center Back and Wide Forward.
- **Vision model** adds aggregated features from combinations of the models outlined in Section 3. We distinguish between several key components: attack and defense surfaces as exemplified by Figure 10a and Figure 10b (e.g., $PC_{jdef}^{0.5}$); the amount of space seen using the *vision map* $V_i$; the ratio of observed occupied space to total occupied space ($V_{rat}$, Formula 10); the ratio of defense to attack seen; the amount of attack (or defense) observed as a percentage of the whole pitch; and finally, the mean values of the aforementioned surface(s) over the time period from $t_{await}^{start}$ until time $t$.

$$V_{rat}^{def} = \frac{\sum PC_{jdef}^{0.5} \odot V_i}{\sum PC_{jdef}^{0.5}} \tag{10}$$

## 4.3. Validation Results

Table 1 shows the results of the four increasingly complex models from our ablation study. We can clearly see the added value of our aggregated vision features, as it improves the regular model performance from 0.744 to 0.788 AUC.

Furthermore, we see that adding traditional VAE features to the baseline model yields no performance increase. Further supporting [12]'s findings that question traditional VAEs as a reliable performance differentiator.

Table 1. An overview of performance metrics for the individual XGBoost models

| Model | AUC | Precision | Recall | F1 |
|---|---|---|---|---|
| Baseline | 0.664 | 0.61 | 0.72 | 0.66 |
| Traditional VEA | 0.654 | 0.60 | 0.74 | 0.66 |
| Regular | 0.744 | 0.69 | **0.78** | **0.74** |
| **Vision** | **0.788** | **0.71** | **0.78** | **0.74** |

## 4.4. Impactful Features

To assess how individual features influence model performance in our validation setup, SHAP values were calculated for all features. The results are shown in Figure 13. Blue (positive) indicates that higher values for these features push predictions toward the positive class (i.e., increasing the imminent pitch value of player $i$ at the end of their *on-ball* phase). In contrast, red indicates that higher values for these features push predictions toward the negative class.

At the top of Figure 13 we notice that seeing a higher amount of defensive occupied space as a total of the entire pitch area averaged over the time during the *awaiting* phase [A], seeing a higher proportion of defensive occupied space compared to attacking occupied space averaged over time during the *awaiting* phase [B], seeing a higher amount of attacking occupied space as a total of the entire pitch area averaged over time during the *awaiting* phase [C] and observing more of the total attack controlled areas averaged over time during the *awaiting* phase [D] are all predictive of an increase in imminent pitch value at the end of the *on ball* phase.

At the bottom of Figure 13 we see that high values for observing more of the total defensive controlled areas averaged over time during the *awaiting* phase [G], and the percentage of instantaneously observed defensive controlled space [H] both are predictive of a decrease in imminent pitch value at the end of the *on ball* phase.

The latter two features potentially show us that observing more defense controlled space could indicate our attacking player is simply in a highly contested area of the pitch, rather than teach us something about the quality of their visual perception. Feature [F] tells us that simply observing more of the field does not lead to better results. And most importantly, features [A], [B], [C] and [D] clearly indicate that players that make higher quality observations (i.e., observe higher quantities of the total occupied space, with a slight focus on identifying defensively controlled space) while they are awaiting a pass, are more likely to increase the quality of the space they occupy at the end of their subsequent on-ball action.

In the middle of Figure 13 we observe that the total number of traditional VEA's - observed at 25 FPS - leading up to time *t* has no impact on our ability to predict if a player is going to gain (or lose) pitch value at the end of the *on-ball* phase.

With the exception of the Wide Forward, position types in general have little to no impact on our ability to predict the change in pitch value at the end of this phase. Furthermore, the percentage of instantaneously observed (i.e., at time *t*) attacking controlled space [E], and the average amount of the total pitch area observed leading up to time *t* in the *awaiting* phase [F], have very little influence on the model outcome.

A rationale for the apparent contradiction in distance to opponent goal and distance to opponent goal line is given in Appendix C.

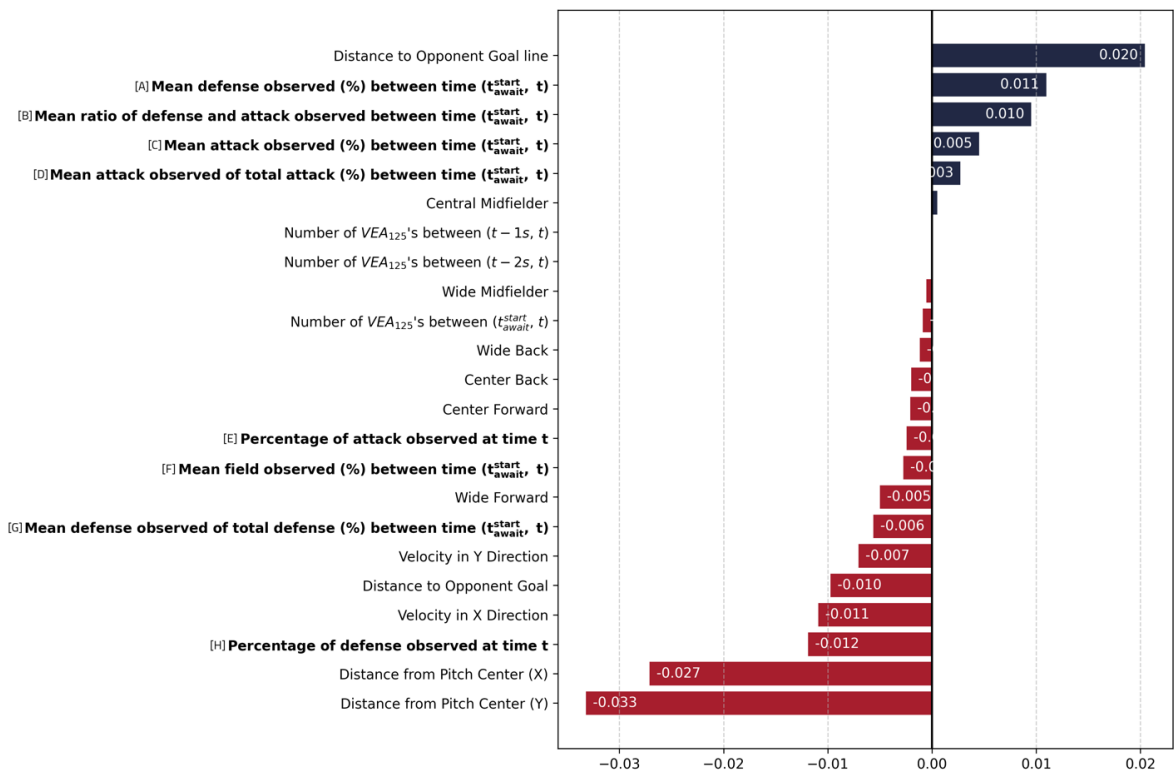

**Fig.13.** The impact and magnitude of individual variables on the XGBoost Classifier that includes all features, using SHAP values. In bold all aggregated vision features created using vision maps labeled A through H. Blue (positive) indicates higher values for variables (e.g., a greater distance to the opponent's goal line) push the prediction toward the positive class (that is., an increase in pitch value at time $t^{end}_{on\_ball}$). Red (negative) indicates increasing values for these variables (e.g., a greater distance from the center of the pitch) push the prediction towards the negative class (that is, a decrease in pitch value at time $t^{end}_{on\_ball}$).

# 5. Conclusion

Within this research we have developed a novel approach to modeling visual perceptual behavior in soccer using a two-dimensional top-down plane using pose-enhanced positional tracking data. This method integrates existing pitch control and pitch value frameworks with a vision map and an occlusion map to quantify controlled and observed pitch value. Within our validation we demonstrate that visual behavior during the awaiting phase shows predictive power for subsequent changes in controlled space after completing a dribbling action. Specifically, players who observe higher quantities of occupied space - particularly defensively controlled areas - while awaiting passes exhibit larger gains in their spatial positioning after subsequent on-ball actions, while traditional visual exploratory action (VEA) counting methods hold zero predictive value for these outcomes. In contrast to VEAs, our approach works independently of player position, eliminates manual annotation requirements, and provides continuous measurements that integrate with existing analytical frameworks such as pitch control, pitch value and SoccerMap. As pose-estimation data becomes increasingly available across invasion sports, this approach has potential applications beyond soccer, including American football and basketball.

# References

[1] Aalbers, B., Van Haaren, J.: Distinguishing between roles of football players in play-by-play match event data. In: Machine Learning and Data Mining for Sports Analytics: 5th International Workshop, MLSA 2018, Co-located with ECML/PKDD 2018, Dublin, Ireland, September 10, 2018, Proceedings 5. pp. 31–41. Springer (2019)

[2] Anzer, G., Arnsmeyer, K., Bauer, P., Bekkers, J., Brefeld, U., Davis, J., Evans, N., Kempe, M., Robertson, S.J., Smith, J.W., et al.: Common data format (cdf): A standardized format for match-data in football (soccer). arXiv preprint arXiv:2505.15820 (2025)

[3] Bassek, M., Rein, R., Weber, H., Memmert, D.: An integrated dataset of spatiotemporal and event data in elite soccer. Scientific Data **12**(1), 195 (2025)

[4] Bauer, P., Anzer, G., Shaw, L.: Putting team formations in association football into context. Journal of sports analytics **9**(1), 39–59 (2023)

[5] Bekkers, J.: Pressing intensity: An intuitive measure for pressing in soccer. arXiv preprint arXiv:2501.04712 (2024)

[6] Bekkers, J.: EFPI: Elastic Formation and Position Identification in Football (Soccer) using Template Matching and Linear Assignment (June 2025), https: //arxiv.org/abs/2506.23843, arXiv preprint arXiv:2506.23843

[7] Bekkers, J.: Wide open gazes: Quantifying visual exploratory behavior in soccer with pose enhanced positional data (2025). https://doi.org/10.6084/m9.figshare.29468036

[8] Bekkers, J., Dabadghao, S.: Flow motifs in soccer: What can passing behavior tell us? Journal of Sports Analytics **5**(4), 299–311 (2019)

[9] Bekkers, J., Sahasrabudhe, A.: A graph neural network deep-dive into successful counterattacks. arXiv preprint arXiv:2411.17450 (2024)

[10] Cao, Z., Simon, T., Wei, S.E., Sheikh, Y.: Realtime multi-person 2d pose estimation using part affinity fields. In: Proceedings of the IEEE Conference on Computer Vision and Pattern Recognition. pp. 7291–7299 (2017)

[11] Casanova, F., Garganta, J., Silva, G., Alves, A., Oliveira, J., Williams, A.M.: Effects of prolonged intermittent exercise on perceptual-cognitive processes. Medicine & Science in Sports & Exercise **45**(8), 1610–1617 (2013)

[12] Caso, S., McGuckian, T.B., van der Kamp, J.: No evidence that visual exploratory activity distinguishes the super elite from elite football players. Science and Medicine in Football pp. 1–9 (2024)

[13] Chalkley, D., Shepherd, J.B., McGuckian, T.B., Pepping, G.J.: Development and validation of a sensor-based algorithm for detecting the visual exploratory actions. IEEE Sensors Letters **2**(2), 1–4 (2018)

[14] Cheng, B., Xiao, B., Wang, J., Shi, H., Huang, T.S., Zhang, L.: Higherhrnet: Scaleaware representation learning for bottom-up human pose estimation. In: Proceedings of the IEEE/CVF Conference on Computer Vision and Pattern Recognition. pp. 5386–5395 (2020)

[15] Decroos, T., Bransen, L., Van Haaren, J., Davis, J.: Actions speak louder than goals: Valuing player actions in soccer. In: Proceedings of the 25th ACM SIGKDD International Conference on Knowledge Discovery and Data Mining. pp. 1851–1861. KDD '19, ACM, New York, NY, USA (2019). https://doi.org/10.1145/3292500.3330758

[16] Eldridge, D., Pulling, C., Robins, M.T.: Visual exploratory activity and resultant behavioural analysis of youth midfield soccer players. Journal of Human Sport and Exercise **8**(3), 560–577 (2013)

[17] Euler, L.: Introductio in analysin infinitorum, vol. 2. MM Bousquet (1748)

[18] Fang, Y., Nakashima, R., Matsumiya, K., Kuriki, I., Shioiri, S.: Eye-head coordination for visual cognitive processing. PloS one **10**(3), e0121035 (2015)

[19] Fernández, J., Bornn, L.: Soccermap: A deep learning architecture for visually-interpretable analysis in soccer. In: Joint European Conference on Machine Learning and Knowledge Discovery in Databases, pp. 491-506 (2020)

[20] Fernández, J., Bornn, L.: Wide open spaces: A statistical technique for measuring space creation in professional soccer. In: Sloan sports analytics conference. vol. 2018 (2018)

[21] Henson, D.: Visual Fields. Oxford University Press, Oxford (1993)

[22] Jordet, G.: Applied cognitive sport psychology in team ball sports: an ecological approach. New Approaches to Sport and Exercise Psychology, eds R. Stelter and KK Roessler (Aachen: Meyer & Meyer Sport) pp. 147–174 (2005)

[23] Jordet, G., Bloomfield, J., Heijmerikx, J.: The hidden foundation of field vision in english premier league (epl) soccer players. In: Proceedings of the MIT sloan sports analytics conference. pp. 1–2 (2013)

[24] Kredel, R., Hernandez, J., Hossner, E.J., Zahno, S.: Eye-tracking technology and the dynamics of natural gaze behavior in sports: an update 2016–2022. Frontiers in Psychology **14**, 1130051 (2023)

[25] Lee, M., Jo, G., Hong, M., Bauer, P., Ko, S.K.: express: Contextual valuation of individual players within pressing situations in soccer. In: Proceedings of the MIT Sloan Sports Analytics Conference. Boston, MA (March 2025)

[26] Maas, T.R.: Monitoring of Visual Exploratory Activity in Professional Football Using a Camera-Based Detection Algorithm. Master's thesis, Eindhoven University of Technology (2025)

[27] McGuckian, T.B., Cole, M.H., Chalkley, D., Jordet, G., Pepping, G.J.: Constraints on visual exploration of youth football players during 11v11 match-play: The influence of playing role, pitch position and phase of play. Journal of Sports Sciences **38**(6), 658–668 (2020)

[28] McGuckian, T. B., Cole, M. H., Jordet, G., Chalkley, D., & Pepping, G. J. : Don't turn blind! The relationship between exploration before ball possession and on-ball performance in association football. Frontiers in psychology, 9, 2520 (2018)

[29] McGuckian, T.B., Cole, M.H., Pepping, G.J.: A systematic review of the technology-based assessment of visual perception and exploration behaviour in association football. Journal of Sports Sciences **36**(8), 861–880 (2018)

[30] Panchuk, D., Vickers, J.N.: Gaze behaviors of goaltenders under spatial–temporal constraints. Human Movement Science **25**(6), 733–752 (2006)

[31] Pokolm, M., Kirchhain, M., Müller, D., Jordet, G., Memmert, D.: Head movement direction in football-a field study on visual scanning activity during the uefa-u17 and -u21 european championship 2019. Journal of Sports Sciences **41**(7), 695–705 (2023)

[32] Power, P., Ruiz, H., Wei, X., Lucey, P.: Not all passes are created equal: Objectively measuring the risk and reward of passes in soccer from tracking data. In: Proceedings of the 23rd ACM SIGKDD international conference on knowledge discovery and data mining. pp. 1605–1613 (2017)

33. Rahimian, P., Van Haaren, J., Abzhanova, T., Toka, L.: Beyond action valuation: A deep reinforcement learning framework for optimizing player decisions in soccer. In: 16th MIT sloan sports analytics conference. vol. 3 (2022)
34. ReSpo.Vision: ReSpo.Vision (2025), https://respo.vision/
35. Robberechts, P., Van Roy, M., Davis, J.: un-xpass: Measuring soccer player's creativity. In: Proceedings of the 29th ACM SIGKDD conference on knowledge discovery and data mining. pp. 4768–4777 (2023)
36. Roca, A., Ford, P.R., McRobert, A.P., Williams, A.M.: Identifying the processes underpinning anticipation and decision-making in a dynamic time-constrained task. Cognitive Processing **12**(3), 301–310 (2011)
37. Spearman, W., Basye, A., Dick, G., Hotovy, R., Pop, P.: Physics-based modeling of pass probabilities in soccer. In: Proceeding of the 11th MIT Sloan Sports Analytics Conference. vol. 1 (2017)
38. Stats Perform: Stats Perform (2025), https://www.statsperform.com/
39. Sun, K., Xiao, B., Liu, D., Wang, J.: Deep high-resolution representation learning for human pose estimation. In: Proceedings of the IEEE/CVF Conference on Computer Vision and Pattern Recognition. pp. 5693–5703 (2019)
40. Taki, T., Hasegawa, J.i.: Visualization of dominant region in team games and its application to teamwork analysis. Computer Graphics International 2000. Proceedings pp. 227–235 (2000)
41. Teranishi, M., Tsutsui, K., Takeda, K., Fujii, K.: Evaluation of creating scoring opportunities for teammates in soccer via trajectory prediction. In: International Workshop on Machine Learning and Data Mining for Sports Analytics. pp. 53–73. Springer (2022)
42. Vaeyens, R., Lenoir, M., Williams, A.M., Mazyn, L., Philippaerts, R.M.: Mechanisms underpinning successful decision making in skilled youth soccer players: An analysis of visual search behaviors. Journal of Motor Behavior **39**(5), 395–408 (2007)
43. Van Roy, M., Cascioli, L., Davis, J.: Etsy: A rule-based approach to event and tracking data synchronization. In: Machine Learning and Data Mining for Sports Analytics ECML/PKDD 2023 Workshop. pp. 11,23. Springer (2023)
44. Vossen, K.: Kloppy: standardizing soccer tracking and event data [github] (2020), https://github.com/PySport/kloppy

# Appendix A. Symbol Definitions

A comprehensive list of symbols in order of first occurrence.

| Symbol | Definition | Section |
|---|---|---|
| $p_i$ | Location of player $i$ | 2.2 |
| $p_j$ | Location of player $j$ | 2.2 |
| $p_b$ | Ball location | 2.2 |
| $\theta_h$ | Head rotation angle | 2.2 |
| $\theta_s$ | Shoulder rotation angle | 2.2 |
| $\odot$ | Element-wise multiplication operator | 3.1 |
| $t$ | Current time | 3.1 |
| $V_\beta$ | Binary field of view for player $i$ | 3.1 |
| $V_p$ | Probabilistic field of view for player $i$ | 3.1 |
| $v_i$ | Speed of player $i$ | 3.1 |
| $R(c_r(v_i))$ | Radial vision component influenced by player speed | 3.1 |
| $A(c_a(v_i))$ | Angular vision component influenced by player speed | 3.1 |
| $c_r$ | Scaling parameter controlling radial vision decay rate | 3.1 |
| $d$ | Distance away from player $i$ | 3.1 |
| $\sigma_r$ | Standard deviation of depth of vision | 3.1 |
| $c_a$ | Scaling parameter controlling angular vision decay rate | 3.1 |
| $\theta_a$ | Angle away from focal point | 3.1 |
| $\sigma_a$ | Standard deviation of angular vision | 3.1 |
| $V_\phi$ | Combined occlusion map for player $i$ | 3.1 |
| $V_{\phi,i,j}$ | Individual occlusion map of player $j$ as perceived by player $i$ | 3.1 |
| $J$ | Set of all other players | 3.1 |
| $V_{o,i,j}$ | Binary mask, the unobstructed view between players $i$ and $j$ | 3.1 |
| $Q_{i,j}$ | Probabilistic ray projected from player $i$ through player $j$ | 3.1 |
| $\alpha$ | Maximum probability of obstruction parameter | 3.1 |
| $\theta_q$ | Angle parameter in occlusion ray formula | 3.1 |
| $\sigma_q$ | Standard deviation for occlusion ray | 3.1 |
| $c_q$ | Scaling parameter for occlusion ray width | 3.1 |
| $\delta_{i,j}$ | Distance between players $i$ and $j$ | 3.1 |
| $\omega_\alpha$ | Apparent angular width of player $j$ as perceived by player $i$ | 3.1 |
| $V$ | Complete vision map of player $i$ | 3.1 |
| $PC$ | Pitch control | 3.2 |
| $c_{in}$ | Scaling parameter for player influence radius $R_i$ | 3.2 |
| $PC^{c_{in}}$ | Imminent pitch control with scaling parameter $c_{in}$ | 3.2 |
| $J^{att}$ | Set of attacking team players | 3.2 |
| $J^{def}$ | Set of defending team players | 3.2 |
| $PC_{j,att}^{0.5}$ | Imminent pitch control of attacking team excluding player $i$ | 3.2 |
| $PC_{j,def}^{0.5}$ | Imminent pitch control of defending team | 3.2 |
| $PC_i^{0.5}$ | Imminent pitch control of attacking player $i$ | 3.2 |
| $R_i$ | Player influence radius | 3.2 |
| $V_l$ | Pitch value surface given ball location | 3.2 |
| $V_\eta$ | Distance to goal pitch normalization surface | 3.2 |
| $\hat{V}_l$ | Normalized pitch value surface | 3.2 |
| $t_{await}^{start}$ | Start time of awaiting phase | 3.2 |
| $t_{await}^{end}$ | End time of awaiting phase | 3.2 |
| $t_{on\_ball}^{end}$ | End time of on-ball phase | 3.2 |
| $V_{rat}$ | Ratio of observed occupied space to total occupied space | 4 |
| $P_v(t)$ | Instantaneous pitch value of space controlled by player $i$ at time $t$ | 4 |
| $p_{rat}$ | Ratio of pitch value increase/decrease | 4 |
| $y$ | Binary label of decrease (0) or increase (1) of pitch value | 4 |
| $\omega_s$ | Assumed shoulder width of a player (0.5m) | C.2 |
| $d_s$ | Assumed torso depth of a player (0.3m) | C.2 |
| $c_k^{local}$ | Local torso position of corner $k$ | C.2 |
| $R(\theta_s)$ | Rotation matrix for player orientation | C.2 |
| $c_k$ | Global torso position of corner $k$ | C.2 |
| $v_k$ | Vector from player $i$ to corner $k$ of player $j$'s body | C.2 |
| $\theta_{a,b}$ | Angle between vectors $a$ and $b$ | C.2 |
| $\theta_{1,3}$ | Angle between opposite corners 1 and 3 of player $j$'s torso | C.2 |
| $\theta_{2,4}$ | Angle between opposite corners 2 and 4 of player $j$'s torso | C.2 |

# Appendix B. Scaling Factors

## Appendix B.1. Peripheral Vision Map Scaling Factors $c_a$ and $c_r$

We define two, empirically determined, linear functions to translate a player's speed ($v_i$) into the scaling factor $c_a$ following Formula 13, and $c_r$ following Formula 14.

$$c_a = \min(0.3\, v_i + 0.2, 0.5) \tag{13}$$

$$c_r = \min(0.25\, v_i + 0.1, 2.6) \tag{14}$$

## Appendix B.2. Occlusion Map Scaling Factor $c_q$

### Apparent Width ($\omega_a$)

The apparent width ($\omega_a$) is an angle (in radians) that represents the angular width of a player $j$'s body - represented as a rectangle - as seen by player $i$ in the two-dimensional top-down plane.

Given position $p_i = [x_i, y_i]$ and position $p_j = [x_j, y_j]$, where player $j$ is represented as a rectangle with width $\omega_s$, depth $d_s$, and orientation $\theta_s$, we calculate the apparent width of the observed body of player $j$ by computing the four points ($k$) of their rectangular torso (in the 2D-plane) in local space with Formula 15a, creating the rotation matrix for the rectangle's orientation using player $j$'s shoulder rotation ($\theta_s$) with Formula 15b, and then calculating the global torso position ($c_k$) for all four points (k) through Formula 15c.

$$c_k^{local} = \left\{ \begin{bmatrix} \frac{\omega_s}{2} \\ \frac{d_s}{2} \end{bmatrix}, \begin{bmatrix} -\frac{\omega_s}{2} \\ \frac{d_s}{2} \end{bmatrix}, \begin{bmatrix} -\frac{\omega_s}{2} \\ -\frac{d_s}{2} \end{bmatrix}, \begin{bmatrix} \frac{\omega_s}{2} \\ -\frac{d_s}{2} \end{bmatrix} \right\} \tag{15a}$$

$$R(\theta_s) = \begin{bmatrix} \cos(\theta_s) & -\sin(\theta_s) \\ \sin(\theta_s) & \cos(\theta_s) \end{bmatrix} \tag{15b}$$

$$c_k = R(\theta_s) \cdot c_k^{local} + p_j, \quad k \in \{1,2,3,4\} \tag{15b}$$

Subsequently, we compute the vectors ($v_k$) from player $i$'s position ($p_i$) to each corner of player $j$'s body ($c_k$) with Formula 15d. Then, the width of player $j$'s body as perceived by player $i$ in the two-dimensional plane is the angle between the lines of sight to the opposite corners of player $j$'s torso as calculated in Formula 15e. Here the angle ($\theta$) between vectors $v_a$ and $v_b$ is calculated for the tow opposite corners [($v_1, v_3$) and ($v_2, v_4$)] of player $j$'s torso. Finally, the apparent width of player $j$'s body as seen from player $i$'s perspective, in radians, is the maximum angle between either pair of opposite corners (Formula 15f).

$$v_k = c_k - p_i, \quad k \in \{1,2,3,4\} \tag{15a}$$

$$\theta_{a,b} = \arccos\left(\frac{v_a \cdot v_b}{\|v_a\| \, \|v_b\|}\right) \tag{15b}$$

$$\omega_\alpha = \max(\theta_{1,3}, \theta_{2,4}) \tag{15b}$$

**Distance Between Player *i* and *j* ($\delta_{i,j}$)**

The distance between player *i* and player *j*, used to appropriately scale $\omega_\alpha$ relative to both player's positions is computed following Formula 16.

$$\delta_{i,j} = \|p_j - p_i\| \tag{16}$$

## Appendix C. Distance To Opponent Goal (Line)

Despite their apparent similarity, "Distance to Opponent Goal Line" and "Distance to Opponent Goal" exert opposite effects on the predicted values. This apparent contradiction is resolved by considering that players positioned at intermediate distances from the goal line but closer to the goal control higher imminent pitch value than those near the goal line but far from goal (e.g., near corner flags). This is further reinforced by the pitch value model (as shown in Figure 11) which shows that wide areas are significantly less valuable, and by the inherit scarcity of samples where a player awaits a pass very close to goal and very close to the goal line (e.g., in the six-yard box) and subsequently executes an on-ball move. Our aggressive filtering approach as discussed in Section 4.2 further enhances this because it eliminates more edge cases.